\pdfoutput=1

\documentclass[11pt]{article}

\usepackage{acl}

\usepackage{times}
\usepackage{latexsym}

\usepackage[T1]{fontenc}

\usepackage[utf8]{inputenc}

\usepackage{microtype}

\usepackage{inconsolata}

\usepackage{graphicx}
\usepackage{subfigure}
\usepackage{booktabs}
\usepackage{multirow}
\usepackage{xurl}

\usepackage{amsmath}
\usepackage{algorithm}
\usepackage{algorithmic}
\usepackage{listings}
\lstdefinestyle{lfonts}{
  basicstyle   = \tiny\ttfamily,
  stringstyle  = \color{purple},
  keywordstyle = \color{blue!60!black}\bfseries,
  commentstyle = \color{olive}\scshape,
}
\lstdefinestyle{lnumbers}{
  numbers     = left,
  numberstyle = \tiny,
  numbersep   = 1em,
  firstnumber = 1,
  stepnumber  = 1,
}
\lstdefinestyle{llayout}{
  breaklines       = true,
  tabsize          = 2,
  columns          = flexible,
}
\lstdefinestyle{lgeometry}{
  xleftmargin      = 20pt,
  xrightmargin     = 0pt,
  frame            = tb,
  framesep         = \fboxsep,
  framexleftmargin = 20pt,
}
\lstdefinestyle{lgeneral}{
  style = lfonts,
  style = lnumbers,
  style = llayout,
  style = lgeometry,
}
\lstdefinestyle{python}{
    language = {Python},
    style    = lgeneral,
}

\title{Training Long-Context LLMs Efficiently via Chunk-wise Optimization}

\author{
  \textbf{Wenhao Li\textsuperscript{1,2}},
  \textbf{Yuxin Zhang\textsuperscript{1}},
  \textbf{Gen Luo\textsuperscript{2}},
  \textbf{Daohai Yu\textsuperscript{1}},
  \textbf{Rongrong Ji\textsuperscript{1}} \\
  \textsuperscript{1}Key Laboratory of Multimedia Trusted Perception and Efficient Computing, \\
  Ministry of Education of China, Xiamen University, 361005, P.R. China \\
  \textsuperscript{2}OpenGVLab, Shanghai AI Laboratory \\
  \small{\textbf{Correspondence:} \href{mailto:rrji@xmu.edu.cn}{rrji@xmu.edu.cn}}
}

\begin{document}
\maketitle
\begin{abstract}
%
While long-context large language models (LLMs) exhibit remarkable document processing capabilities, their prohibitively high training costs often hinder customized applications. To mitigate this issue, we propose \textit{Sequential Chunk-wise Optimization} (SeCO), a memory-efficient training paradigm that partitions lengthy inputs into manageable chunks. Each chunk independently constructs its computational graph and performs localized backpropagation, ensuring that only one chunk's forward activations are stored in memory.
Building on SeCO, we further introduce \textit{Sparse Chunk-wise Optimization} (SpaCO), which reduces computational overhead by selectively propagating gradients to specific chunks and incorporates a carefully designed compensation factor to ensure unbiased gradient estimation.
SpaCO decouples the computational cost of backpropagation from the context length, enabling training time to gradually converge to inference time as sequences become longer.
Implemented as lightweight training wrappers, both SeCO and SpaCO offer substantial practical benefits. 
For example, when fine-tuning an 8B model with LoRA on a single RTX 3090 GPU, SeCO expands maximum sequence length from 1K to 16K tokens, while SpaCO demonstrates accelerated training speed---achieving up to 3× faster than SeCO under the same experimental setup.
These innovations provide new insights into optimizing long-context models, making them more accessible for practical applications.
We have open-sourced the code at \href{https://github.com/wenhaoli-xmu/seco}{here}.
\end{abstract}

\begin{figure*}[ht]
\centering
\includegraphics[width=0.95\textwidth]{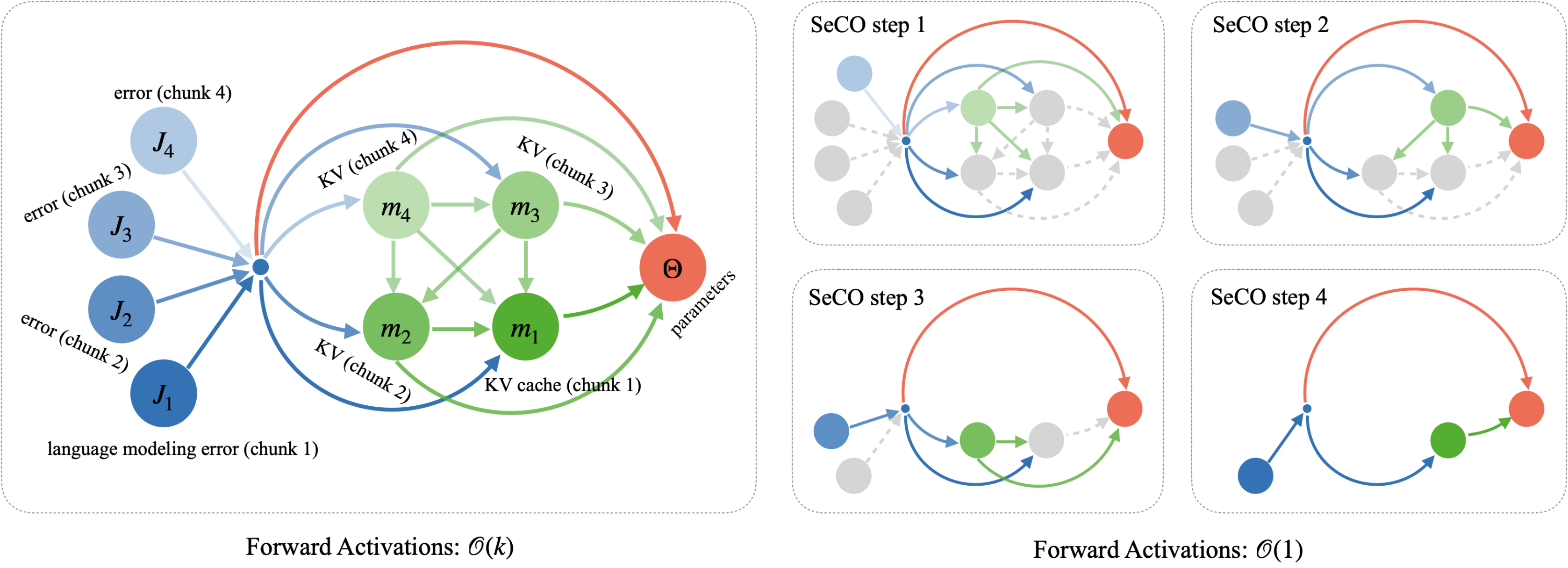}
\caption{(\textit{Left}) Computational graph for chunk-wise optimization with $k=4$ chunks. The dense connections among KV caches (green arrows) complicate memory management, leading popular training frameworks~\citep{deepspeed,accelerate} to rely on end-to-end parallel training. (\textit{Right}) By analyzing the topology of this graph, we propose SeCO, a bootstrapping method leveraging gradient checkpointing along the sequence dimension. SeCO ensures that only the computational graph of a single chunk is stored at any time.}
\label{fig:seco-pipe}
\end{figure*}
\section{Introduction}
\label{sec:intro}

Recent advancements in long-context LLMs~\citep{chen2024longlora,10.5555/3692070.3692130,peng2024yarn,zhao2024large} have demonstrated unprecedented capabilities in processing lengthy documents, offering superior retrieval quality compared to retrieval-augmented generation (RAG) approaches~\citep{Liu_LlamaIndex_2022}, making them particularly valuable for commercial applications requiring nuanced document understanding.

However, fine-tuning these models faces significant resource challenges: \emph{(i) Time Overhead:} The quadratic scaling of attention mechanisms leads to prohibitive training time~\citep{harm-de-vries}. \emph{(ii) Memory Constraints:} Despite optimizations like FlashAttention~\citep{dao2024flashattention}, the storage requirements for forward activations still increases linearly with sequence length, quickly depleting GPU memory. 
As a result, fine-tuning 8B models~\citep{grattafiori2024llama3herdmodels} with LoRA~\citep{hu2022lora} on a single RTX 3090 GPU is limited to sequences of only 1K tokens.

%
Existing architectural modifications, exemplified by LongLoRA's $S^2$-attention~\citep{chen2024longlora}, aim to alleviate these issues by reducing computational overhead to sub-quadratic through attention approximation.
However, these methods incur gradient accuracy compromises while offering limited resource savings,\footnote{As shown in Figure~1 (\textit{Mid}) of the LongLoRA paper, the proposed method exhibits memory scaling patterns similar to those of full fine-tuning baseline, achieving only about a 2-fold extension in sequence length.} motivating our exploration of alternative efficiency improvements strategies.

We introduce \textit{Sequential Chunk-wise Optimization} (SeCO), a novel training method that preserves exact gradients while dramatically reducing memory consumption.
The key innovation of SeCO is the application of gradient checkpointing~\citep{gdckpt,chen2016trainingdeepnetssublinear} along the sequence dimension using chunk-level checkpoints.
This approach represents a fundamental departure from traditional gradient checkpointing, which typically employs a fixed number of checkpoints for static layer-wise or block-wise partitioning of the computational graph.
Unlike these conventional techniques, where memory requirements for forward activations scale linearly with sequence length, SeCO maintains a constant overhead for forward activations, regardless of sequence length.
This innovation achieves order-of-magnitude memory reduction while maintaining manageable training time overhead, establishing it as an efficient solution for fine-tuning long-context LLMs under resource-constrained conditions.

While SeCO successfully mitigates memory constraints in long-context LLM fine-tuning, it maintains computational overhead comparable to naive parallel training.
This computational burden significantly undermines its applicability for processing extended sequences, thereby limiting its practical utility.
To address this limitation, we propose \textit{Sparse Chunk-wise Optimization} (SpaCO), an enhanced variant of SeCO that achieves substantial computational savings.
SpaCO preserves the integrity of forward propagation while implementing selective backpropagation through a fixed subset of chunks.
This modification decouples computational cost from sequence length during gradient computation.
Our theoretical framework reveals a crucial architectural insight: The gradient chain length between key-value (KV) cache chunks exhibits inherent boundedness determined by model depth (as also noted in~\citealp{tsfm-xl}).
This fundamental property enables SpaCO to employ randomized chunk sampling while preserving unbiased gradient estimation, achieving significant computational reduction without compromising theoretical guarantees (for more detailed explanation, please refer to Section~\ref{sec:spaco}).

Empirical evaluations highlight the substantial practical advantages of SeCO and SpaCO:
\begin{itemize}
\item \textbf{Scalability}: 
The memory overhead for SeCO and SpaCO scales minimally with increasing sequence length, as the only contributing factor is the storage of the KV cache.
Moreover, SpaCO's training time converges to inference time as the sequence length expands, demonstrating efficient computational scaling.

\item \textbf{Performance}: Although SpaCO does not compute exact gradients like SeCO, it incurs only a small performance gap. Specifically, at a sparsity ratio of $1/8$, the language modeling error increases by less than 0.1 compared to exact gradient training.
\end{itemize}

Our contributions are threefold:
\begin{itemize}
\item A memory efficient training paradigm (SeCO) that enables long-context fine-tuning through sequence dimensional gradient checkpointing.
\item A computation-efficient extension (SpaCO) leveraging sparsification, with theoretical guarantees of unbiased gradient estimation.
\item Open-source implementations that achieve up to an order of magnitude training sequence length improvements on consumer hardware.
\end{itemize}

\begin{figure}[h]
\centering
\includegraphics[width=0.95\linewidth]{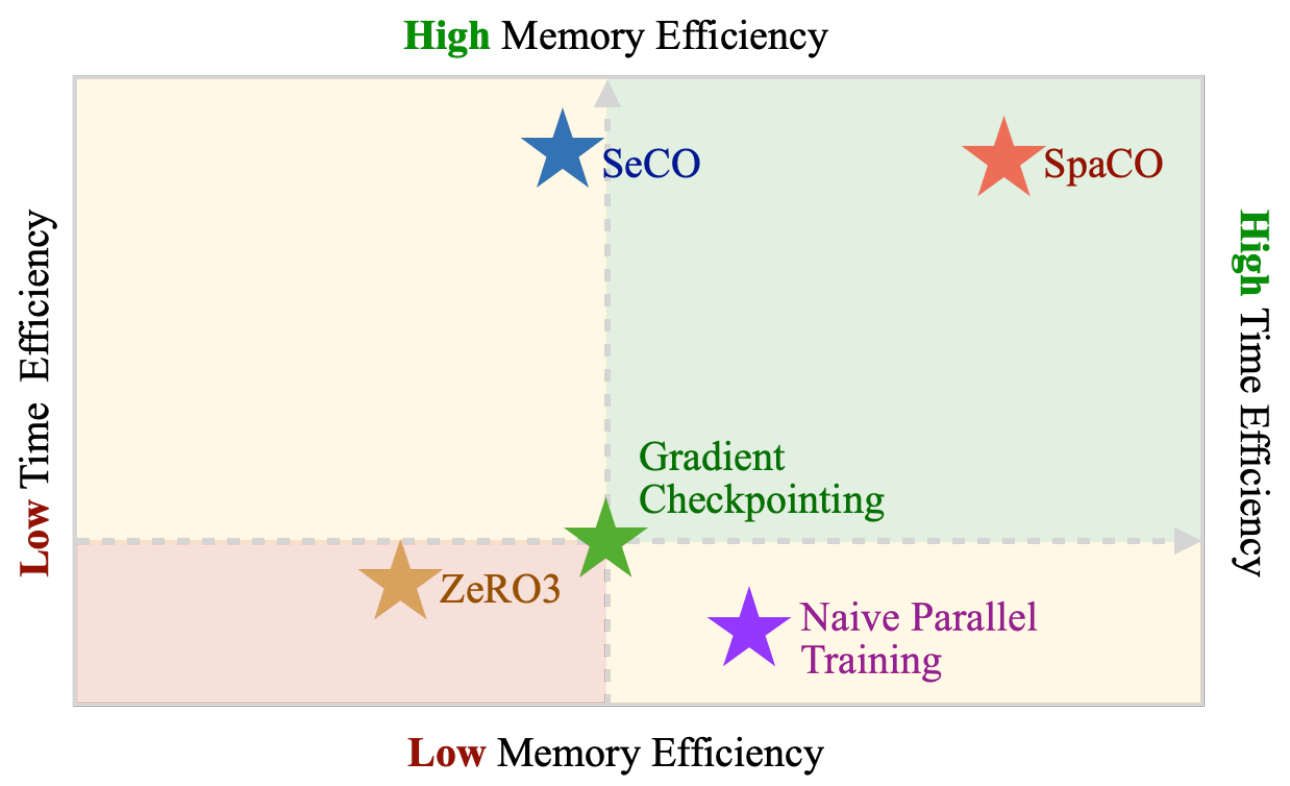}
\caption{Qualitative comparison of different methods. \textit{(i)} SeCO achieves significant memory reduction while maintaining time efficiency comparable to layer-level gradient checkpointing.
\textit{(ii)} Building upon SeCO, SpaCO significantly reduces computational overhead, making the training time converges to inference time as the sequence length expands.}
\end{figure}
\section{Related Works}

\textbf{Long-Context LLMs.}
Efforts to extend the context window of LLMs primarily rely on augmenting positional embeddings and applying limited post-training to adapt models pre-trained on shorter contexts~\citep{chen2024longlora,peng2024yarn}.
While these methods are effective, the inherent quadratic computational complexity of attention mechanisms renders long-context training prohibitively expensive~\citep{harm-de-vries,hu2024longreciperecipeefficientlong}.
Recent works, such as LongLoRA~\citep{chen2024longlora}, address this issue using $S^2$-attention, which achieves linear computation scalability. However, its architectural modification introduces biased gradient computation.

\vspace{0.3\baselineskip} \noindent
\noindent\textbf{Gradient Checkpointing.}
%
Gradient checkpointing techniques~\citep{chen2016trainingdeepnetssublinear,gdckpt} optimize memory consumption by recomputing activations during backpropagation rather of storing them.
In Transformer~\citep{NIPS2017_3f5ee243} architectures, conventional layer-level checkpointing offers limited benefits for long sequences due to the static partitioning of the computational graph.
Gradient checkpointing applied along the sequence dimension enables maintaining a constant memory footprint for storing forward activations, presenting substantial advantages. 
Nevertheless, current implementations in mainstream deep learning frameworks~\citep{torch,tensorflow} are primarily designed for checkpointing within individual forward pass, lacking the capability to handle concatenated computational graphs that emerge across multiple iterative processes.

\vspace{0.3\baselineskip} \noindent
\noindent\textbf{Gradient Estimation.}
The concept of approximate gradient predates modern deep learning, exemplified by stochastic gradient descent's use of mini-batch~\citep{NIPS2007_0d3180d6}.
SpaCO introduces a novel paradigm that aligns with the philosophy of SGD: by utilizing a limited number of gradient propagation pathways to estimate the underlying true gradient, it significantly reduces computational overhead while maintaining acceptable performance.

\section{Preliminary}
When processing large amounts of data all at once, GPU threads can become saturated, causing parallel processing time to scale linearly with data size, offering no advantage over sequential processing while increasing memory usage.
To address this, efficient LLM serving frameworks such as vLLM~\citep{kwon2023efficient} and FlashInfer~\citep{ye2025flashinfer} adopt a chunk pre-filling strategy, splitting long contexts into smaller chunks and processing them sequentially.
We extend this idea from LLM inference to LLM training.

\vspace{0.3\baselineskip} \noindent
\noindent\textbf{Computational Graph.} For an input sequence $X$ partitioned into $k$ chunks $\{x_j\}_{j=1}^k$, let $m_j$ denote the KV cache and $J_j$ the error component for chunk $j$.
The model $f$ with parameter $\Theta$ processes chunks sequentially:
\begin{equation}
(J_j, m_j)=f(x_j ; m_1,m_2,...,m_{j-1}; \Theta).
\label{con:fwd}
\end{equation}
The parameter gradient combines direct and indirect contributions through KV cache:
\begin{equation}
\nabla_\Theta J_j = \underbrace{\frac{\partial J_j}{\partial \Theta}}_{\text{Direct term}} + \sum_{i=1}^j \underbrace{\frac{\text{d} J_j}{\text{d} m_i} \frac{\partial m_i}{\partial \Theta}}_{\text{Indirect contributions}}.
\label{con:grad-1 }
\end{equation}
Due to the iterative nature of $f(\cdot)$, the computation of $\text{d} J_j / \text{d} m_i$ involves nested dependencies:
\begin{equation}
\begin{aligned}
\frac{\text{d} J_j}{\text{d} m_i} \frac{\partial m_i}{\partial \Theta}
&= \frac{\partial J_j}{\partial m_i}\frac{\partial m_i}{\partial \Theta} \\
&+ \sum_{i<t_1\le j}
\frac{\partial J_j}{\partial m_{t_1}}
\frac{\partial m_{t_1}}{\partial m_i}
\frac{\partial m_i}{\partial \Theta} \\
&+ \sum_{i< t_1 < t_2 \le j}
\frac{\partial J_j}{\partial m_{t_2}}
\frac{\partial m_{t_2}}{\partial m_{t_1}}
\frac{\partial m_{t_1}}{\partial m_i}
\frac{\partial m_i}{\partial \Theta} \\
&+\dots .
\end{aligned}
\label{con:grad-2}
\end{equation}
Although Eq.\,(\ref{con:grad-2}) appears complex, it fundamentally demonstrates that gradients propagate through all possible multi-hop paths among $\{m_t\}_{t=i}^j$. 
To visualize this process, we present partial derivatives $\partial\bigtriangleup/\partial\,\bigcirc$ as directed edges from $\bigtriangleup$ to $\bigcirc$, forming the complete computational graph of gradient propagation from $\{J_j\}_{j=1}^k$ to $\Theta$ as illustrated in Figure\,\ref{fig:seco-pipe}.
This graph explicitly captures the computational dependencies for chunk-wise optimization.

\begin{figure}[h]
\centering
\includegraphics[width=0.95\linewidth]{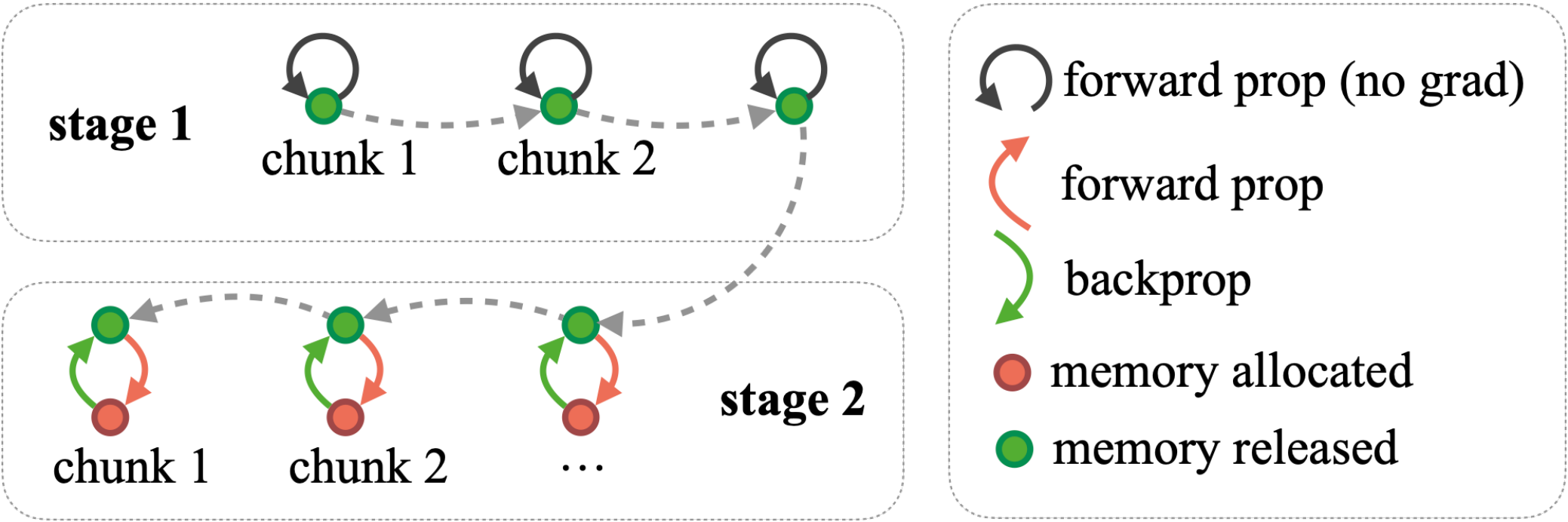}
\caption{A visualized illustration of Algorithm\,\ref{alg:1}. \textit{Stage 1} corresponds to the first for-loop, generating KV caches for all data chunks through inference-mode, serving as checkpoints. \textit{Stage 2} corresponds to the second for-loop, where the computational graph is constructed and localized backpropagation is performed.}
\label{fig:seco-demo}
\end{figure}
\vspace{0.3\baselineskip} \noindent
\noindent\textbf{Gradient Checkpointing.}
Gradient checkpointing trades computational time for reduced memory usage.
The fundamental principle guiding checkpoint placement requires that the complete subsequent computational graph must be reconstructible using only these designed checkpoints and existing leaf nodes.
As formalized in Eq.\,(\ref{con:fwd}), the output of chunk-$j$ is uniquely determined by two components: \textit{(i)} the preceding KV cache sequence $\{m_i\}_{i=1}^{j-1}$ and \textit{(ii)} the model parameters $\Theta$ (leaf nodes).
By storing these KV caches, we enable the complete reconstruction of any subsequent chunk's output during backpropagation, making them ideal checkpoint candidates.
During the forward propagation, only the checkpointed KV caches need to be computed and stored, eliminating the need to retain intermediate activations.

\section{Sequential Chunk-wise Optimization}
SeCO is a plain version of this chunk-wise optimization that does not save any computation and obtains exact gradients (verified in Appendix\,\ref{app:more}).

\subsection{Methodology} 
During forward propagation, we compute all chunks sequentially in inference mode to generate corresponding KV caches $\{m_1^\prime, m_2^\prime,\dots, m_k^\prime\}$, where prime notation distinguishes inference-generated caches from training-phase counterparts.

For backpropagation, the computational graph topology in Figure\,\ref{fig:seco-pipe} dictates a sequential reverse-order reconstruction strategy.
For chunk $j$:
\begin{enumerate}
    \item Reconstruct computational graph using Eq.\,(\ref{con:fwd}) to compute error $J_j$ and KV cache $m_j$.
    \item Transfer gradients from $m_j^\prime$ to $m_j$.
    \item Backpropagate $J_j$ and $m_j$ to accumulate gradients for model parameters and preceding checkpoints $m_1^\prime,\dots,m_{j-1}^\prime$.
\end{enumerate}

After processing all chunks, accumulated parameter gradients match those from naive parallel training modulo numerical precision. Implementation details follow Algorithm \ref{alg:1}, and the corresponding visualization is presented in Figure\,\ref{fig:seco-demo}.

\subsection{Efficiency}
We analyze the theoretical efficiency of SeCO, in terms of computation and storage.

\vspace{0.3\baselineskip} \noindent
\noindent
\textbf{Memory Savings.} By reconstructing at most one chunk's computational graph at any given time, SeCO effectively prevents forward activations from scaling linearly with sequence length.
This design reduces the memory requirements for storing forward activations by a factor of $k$.
However, it is important to note that SeCO does not optimize fixed memory components such as optimizer states and model parameters, nor does it alleviate the memory overhead of the KV cache.

\vspace{0.3\baselineskip} \noindent
\noindent
\textbf{Computational Overhead.}
SeCO introduces two primary sources of computational overhead: \textit{(i)} additional recomputation during backpropagation, and \textit{(ii)} frequent kernel launches for small-scale tensor operations.

For the first component, since backpropagation typically requires approximately twice the FLOPs of forward propagation~\citep{ds-profiler,ml-notes}, the subgraph reconstruction introduces an estimated 33\% computational overhead.

Regarding the second component, modern GPUs like the RTX 3090 contain fixed computational resources (82 streaming multiprocessors with 128 cores each).
When using sufficiently large chunk sizes, these resources can achieve near-saturation utilization.
Experimental results demonstrate that increasing chunk sizes beyond 128 yields diminishing returns, with only marginal reductions in computational time observed.
\begin{algorithm}
\caption{Sequential Chunk-wise Optimization}
\label{alg:1}
\begin{algorithmic}[1]
\REQUIRE Model $f$, data $X = \{x_1, x_2, \ldots, x_k\}$, parameters $\Theta$
\ENSURE $\nabla_{\Theta}$
\FOR{$i = 1$ to $k$}
    \STATE $m_i^\prime \gets f(x_i; \{m_j^\prime\}_{j=1}^{i-1};\Theta)$ 
\ENDFOR

\FOR{$i = k$ to $1$}
    \STATE $J_i,m_i \gets f(x_i; \{m_j^\prime\}_{j=1}^{i-1};\Theta)$    \STATE $m_i\texttt{.grad} \gets m_i^\prime\texttt{.grad}$
    \STATE $\text{backprop}(J_i)$
\ENDFOR
\end{algorithmic}
\end{algorithm}

\begin{figure}[ht]
\centering
\includegraphics[width=0.95\linewidth]{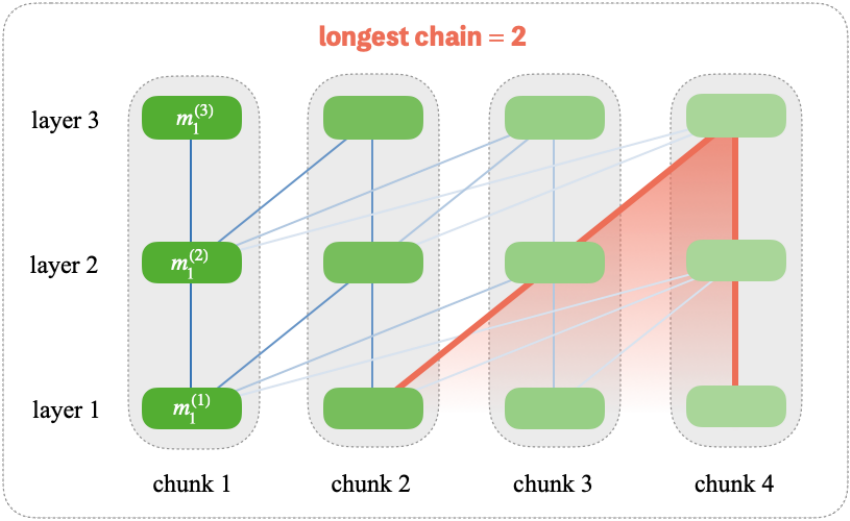}
\caption{In the Transformer architecture, the gradient flow traverses through at most a number of KV caches equal to the layer depth. This observation was also highlighted in Transformer-XL~\citep{tsfm-xl}.}
\label{fig:spaco-longest}
\end{figure}
\section{Sparse Chunk-wise Optimization}
\label{sec:spaco}
While SeCO reduces memory consumption, it introduces additional computational overhead, further prolonging the already time-consuming training process.
This limitation hinders its ability to handle ultra-long sequences efficiently.
To alleviate this issue, we propose SpaCO, an improvement over SeCO. 
By introducing sparsification in backpropagation, SpaCO significantly accelerates training while maintaining memory efficiency.

The key insight stems from the observation that the checkpoints $\{m_1^\prime, \dots, m_k^\prime\}$ enable independent computational graph construction for individual chunk through Eq.\,(\ref{con:fwd}).
Capitalizing on this, SpaCO implements a stochastic backpropagation scheme that randomly selects a subset of chunks for gradient computation during each training iteration.

This sparsification, however, poses the risk of biased gradient estimation. In extreme cases where only one chunk is selected, the gradient flow between chunks is disrupted---akin to non-overlapping chunked attention mechanisms, which constrain the model to local dependencies.

One might hypothesize that dense gradient propagation is essential for learning global patterns, given that the longest gradient chain spans all KV cache chunks.
However, our theoretical analysis reveals that this is not necessary.

\vspace{0.3\baselineskip}
\noindent
\textbf{The Longest Gradient Chain.}
In the computational graph shown in Figure\,\ref{fig:seco-pipe}, the longest gradient chain is:
\begin{equation}
\frac{\partial J_4}{\partial m_4}\cdot
\frac{\partial m_4}{\partial m_3}\cdot
\frac{\partial m_3}{\partial m_2}\cdot
\frac{\partial m_2}{\partial m_1}\cdot
\frac{\partial m_1}{\partial \Theta},
\end{equation}
which spans all chunks. Omitting any chunk would break this chain, leading to biased gradient estimation.
However, in the Transformer~\citep{NIPS2017_3f5ee243} architecture, KV cache chunks within the same layer are independent and computed in parallel.
As a result, errors propagate from one KV cache chunk to another only between adjacent layers~\citep{tsfm-xl}, as shown in Figure\,\ref{fig:spaco-longest}. Thus, the maximum gradient chain length is bounded by the number of layers.
Theoretically, unbiased gradient estimation is achievable if the number of selected chunks meets the number of layers, ensuring sufficient coverage.

\begin{figure}[h]
\centering
\hfill
\begin{minipage}[b]{0.475\linewidth}
    \centering
    \subfigure[Dense graph ($k$ chunks).]{\includegraphics[width=\linewidth]{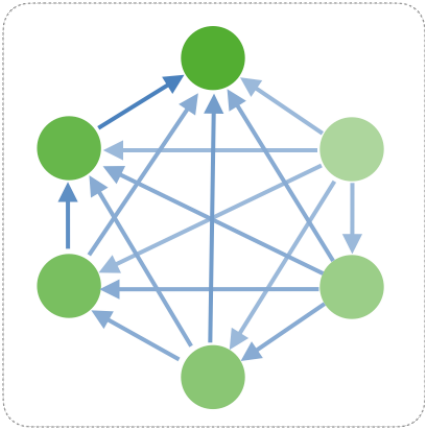}
    \label{fig:spaco-dense}}
    \vfill 
\end{minipage}
\hfill
\begin{minipage}[b]{0.475\linewidth}
    \centering
    \subfigure[Sparse graph ($t$ chunks).]{\includegraphics[width=\linewidth]{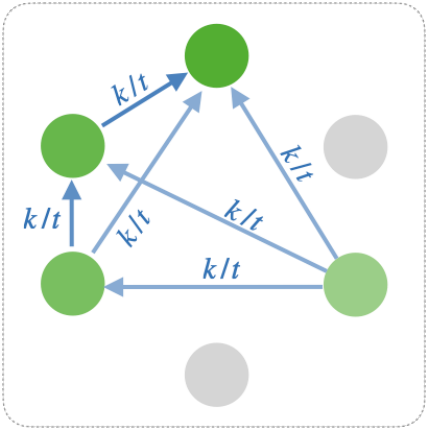}
    \label{fig:spaco-sparse}}
    \vfill 
\end{minipage}
\hfill
\caption{SpaCO sparsifies the gradient flow among KV caches. (\textit{Left}) The original graph. $k=6$. (\textit{Right}) Only the gradient flow between $t=4$ chunks is retained. By adding a factor $k/t$ to each path, the gradient computed from this sparse graph remains an unbiased estimate.}
\label{fig:spaco-graph}
\end{figure}

\vspace{0.3\baselineskip}
\noindent
\textbf{Challenges in Unbiased Estimation.}
While bounded gradient chain length suggests theoretical feasibility, practical implementation faces significant hurdles.
Consider a DAG with $n$ nodes and $n(n-1)/2$ edges, as shown in Figure\,\ref{fig:spaco-graph}\,(\textit{Left}). The number of $p$-length paths follows combinatorial principles:
\begin{equation}
d_p = \binom{n}{p+1} = \frac{n!}{(p+1)!(n-p-1)!}.
\end{equation}
Let superscripts \texttt{d} and \texttt{s} denote dense ($k$ chunks) versus sparse ($t$ chunks) configurations respectively. The path count ratio between these two configurations exhibits:
\begin{equation}
\frac{d_p^\texttt{d}}{d_p^\texttt{s}} = \frac{k(k-1)(k-2)\cdots(k-p)}{t(t-1)(t-2)\cdots(t-p)},
\end{equation}
for $t \gg p$, this simplifies to:
\begin{equation}
\frac{d_p^\texttt{d}}{d_p^\texttt{s}} \approx \left(\frac{k}{t}\right)^{p+1}.
\end{equation}
This exponentially decaying pattern highlights a crucial insight: graph sparsification disproportionately weakens longer gradient chains.
Since removing a node multiplicatively affects all paths passing through it, longer chains suffer a greater cumulative impact.
Consequently, naive sparsification introduces systematic bias.

To ensure unbiased estimation, we must incorporate compensation mechanisms that counteracts this attenuation.
A viable solution is to strategically apply scaling factors to the preserved paths, effectively rebalancing gradient contributions across different chain lengths.

\vspace{0.3\baselineskip}
\noindent
\textbf{Compensation Factor.}
To analyze gradient propagation under sparsification, we consider all gradient chains of length $p$ in Eq.\,(\ref{con:grad-2}), denoted by $\mathbf{z}_p$:
\begin{equation}
\mathbf{z}_p=\sum_{i<t_1<...<t_{p-1}\le j}
\frac{\partial J_j}{\partial m_{t_{p-1}}}\cdot
\frac{\partial m_{t_{p-1}}}{\partial m_{t_{p-2}}}\cdot...\cdot
\frac{\partial m_{i}}{\partial \Theta}.
\label{con:zp}
\end{equation}
A crucial observation is that any gradient chain in $\mathbf{z}_p$ necessitates the sampling of all its constituent chunks.
The survival probability of such a chain under $t$-out-of-$k$ sparse sampling can be derived as $(t/k)^p$ based on the following reasoning:
\begin{itemize}
    \item The initial term $\partial J_j / \partial m_{t_{p-1}}$ survives with probability $t/k$.
    \item Given that the previous term survives, each subsequent term (except the last) also survives with probability $t/k$.
\end{itemize}
Using this survival probability, we can express the expected value of $\mathbf{z}_p$ after sparsification, denoted as $\bar{\mathbf{z}}_p$:
\begin{equation}
\bar{\mathbf{z}}_p=\left(\frac{t}{k}\right)^p \mathbf{z}_p + \left(1-\frac{t^p}{k^p}\right) 0=\left(\frac{t}{k}\right)^p \mathbf{z}_p.
\end{equation}
Given the complete gradient $\mathbf{Z}=\sum_{p=1}^{\infty}\mathbf{z}_p$ from Eq.\,(\ref{con:grad-2}), its expectation after sparsification becomes:
\begin{equation}
\bar{\mathbf{Z}}=\frac{t}{k} \mathbf{z}_1 + \left(\frac{t}{k}\right)^2 \mathbf{z}_2 + \left(\frac{t}{k}\right)^3 \mathbf{z}_3 + ...
\end{equation}
To achieve unbiased gradient estimation ($\bar{\mathbf{Z}}=\mathbf{Z}$), each gradient chain $\mathbf{z}_p$ requires compensation by factor $(k/t)^p$. This multiplicative scaling counteracts the exponential decay induced by sparsity.

\begin{algorithm}
\caption{Sparse Chunk-wise Optimization}
\label{alg:2}
\begin{algorithmic}[1]
\REQUIRE Model $f$, data $X = \{x_1, x_2, \ldots, x_k\}$, parameters $\Theta$, fixed budget $t$
\ENSURE $\nabla_{\Theta}$
\FOR{$i = 1$ to $k$}
    \STATE $m_i^\prime \gets f(x_i; \{m_j^\prime\}_{j=1}^{i-1};\Theta)$
\ENDFOR

\STATE Randomly select $t$ distinct indices from $\{1,...,k\}$, denoted as $\mathcal{I}$.

\FOR{$i$ in $\mathcal{I}$}
    \STATE $J_i,m_i \gets f(x_i; \{m_j^\prime\}_{j=1}^{i-1};\Theta)$
    
    \STATE $m_i\texttt{.grad} \gets \left(\frac{k}{t}\right)\cdot m_i^\prime\texttt{.grad}\cdot$
    \STATE $\text{backprop}(J_i)$
\ENDFOR
\end{algorithmic}
\end{algorithm}

\vspace{0.3\baselineskip}
\noindent
\textbf{Implementation.}
The compensation factor is implemented through modifying backpropagation, which occurs during gradient computation: When calculating $\partial m_i/\partial \{m_j^\prime \}_{j=1}^{i-1}$, we scale the gradient by $k/t$.
Through the nested structure of $f(\cdot)$, this creates compound compensation where each $p$-length chain automatically accumulates $(k/t)^p$ scaling through successive operations.

Figure\,\ref{fig:spaco-graph}\,(\textit{Right}) illustrates this dynamic scaling mechanism, with full implementation details provided in Algorithm\,\ref{alg:2}.
This approach enhances computational efficiency while maintaining the statistical accuracy of full backpropagation.

\begin{figure*}[h]
\includegraphics[width=\linewidth]{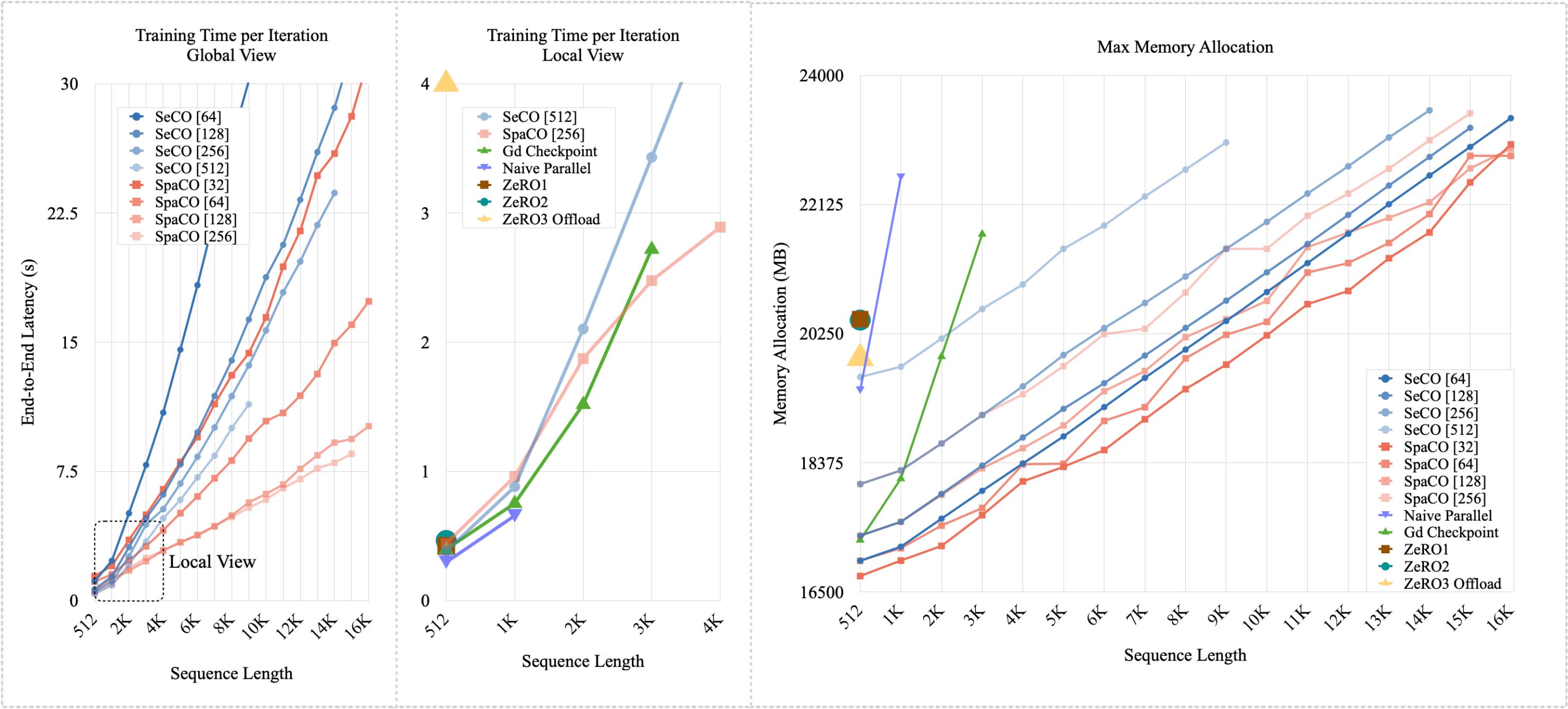}
\caption{(\textit{Left}) Compared to SeCO, SpaCO achieves lower training time with more favorable linear-like scaling as the sequence length increases. (\textit{Middle}) A zoomed-in view of the left panel shows that SeCO only incurs $\sim$30\% additional time overhead compared to naive parallel training, significantly outperforming ZeRO3 offload which suffers from GPU-CPU communication bottlenecks and demonstrates approximately 10× slower training speed. (\textit{Right}) SeCO and SpaCO exhibit superior memory efficiency, achieving more than 4× memory reduction compared to standard gradient checkpointing and an order-of-magnitude improvement over naive parallel training.}
\label{fig:time-memory}
\end{figure*}
\section{Experiment}
We designed experiments to address two key questions:
\begin{itemize}
\item How do SeCO and SpaCO compare to mainstream training methods in time and memory efficiency?
\item Can SpaCO provide reliable gradient estimation and maintain competitive performance compared to exact gradient training?
\end{itemize}
The following sections present our comprehensive analysis. Additional experiments, including the verification of gradient computation accuracy for SeCO, are presented in Appendix\,\ref{app:more}.

\subsection{Experimental Setup}
Our experiments utilize the LLaMA3-8B~\citep{grattafiori2024llama3herdmodels} as the base model, implementing LoRA~\citep{hu2022lora} fine-tuning with hyperparameters $r=8$ and $\alpha=16$.
The training datasets comprises 1,000 instances sampled from the PG19 training split~\citep{pg19}, with sequence lengths truncated to 16K tokens.

\subsection{Baseline Methods}
We compare our methods against three mainstream training paradigms: DeepSpeed~\citep{deepspeed}, conventional layer-level gradient checkpointing, and naive parallel training, all of which leveraging FlashAttention~\citep{dao2024flashattention}.

\vspace{0.3\baselineskip}
\noindent
\textbf{DeepSpeed.}
DeepSpeed~\citep{deepspeed} is a high-performance distributed training framework based on \textit{Fully Sharded Data Parallel} (FSDP). It provides three optimization levels, ZeRO1/2/3, which progressively reduce memory consumption by distributing optimizer states, gradients, and parameter across multiple GPUs.
ZeRO3 offload further extends ZeRO3 by offloading these components to CPU for additional memory savings.
We evaluate DeepSpeed on 8 RTX 3090 GPUs with default ZeRO1/2 configurations and a custom ZeRO3 setup (Appendix\,\ref{app:zero3}).
Notably, FSDP offers limited benefits in parameter-efficient scenarios, where its advantages may not fully emerge.

\vspace{0.3\baselineskip}
\noindent
\textbf{Gradient Checkpointing.}
A standard gradient checkpointing implementation on a single RTX 3090 GPU, placing checkpoints at each LLM layer's input hidden states to reduce memory usage.

\vspace{0.3\baselineskip}
\noindent
\textbf{Naive Parallel Training.}
A baseline implementation on a single RTX 3090 GPU, utilizing no memory-efficient techniques except for FlashAttention. This serves as a reference for time efficiency.

\subsection{Efficiency Analysis}
We evaluate time and memory efficiency of SeCO and SpaCO implemented on a single RTX 3090 GPU with varying sequence lengths.
\begin{itemize}
    \item \textbf{Configuration:} For SeCO, we evaluate chunk sizes \{64,\,128,\,256,\,512\}. For SpaCO, we use a chunk budget of $t=8$ and test with chunk sizes \{32,\,64,\,128,\,256\}.\footnote{We observe that $t=8$ already achieves satisfactory performance in practice.}
    \item \textbf{Measurement Protocol:} Peak memory usage recorded via PyTorch's memory profiler, and the minimum end-to-end iteration time among the first 10 iterations is reported.
\end{itemize}
Our findings are concluded in Figure\,\ref{fig:time-memory}.

\vspace{0.3\baselineskip}
\noindent
\textbf{Practical Guidance.}
Based on our findings, we recommend maximizing the chunk size within the available memory limits.
This accelerates training while maintains the same memory scalability.

\begin{figure*}[h]
\includegraphics[width=\linewidth]{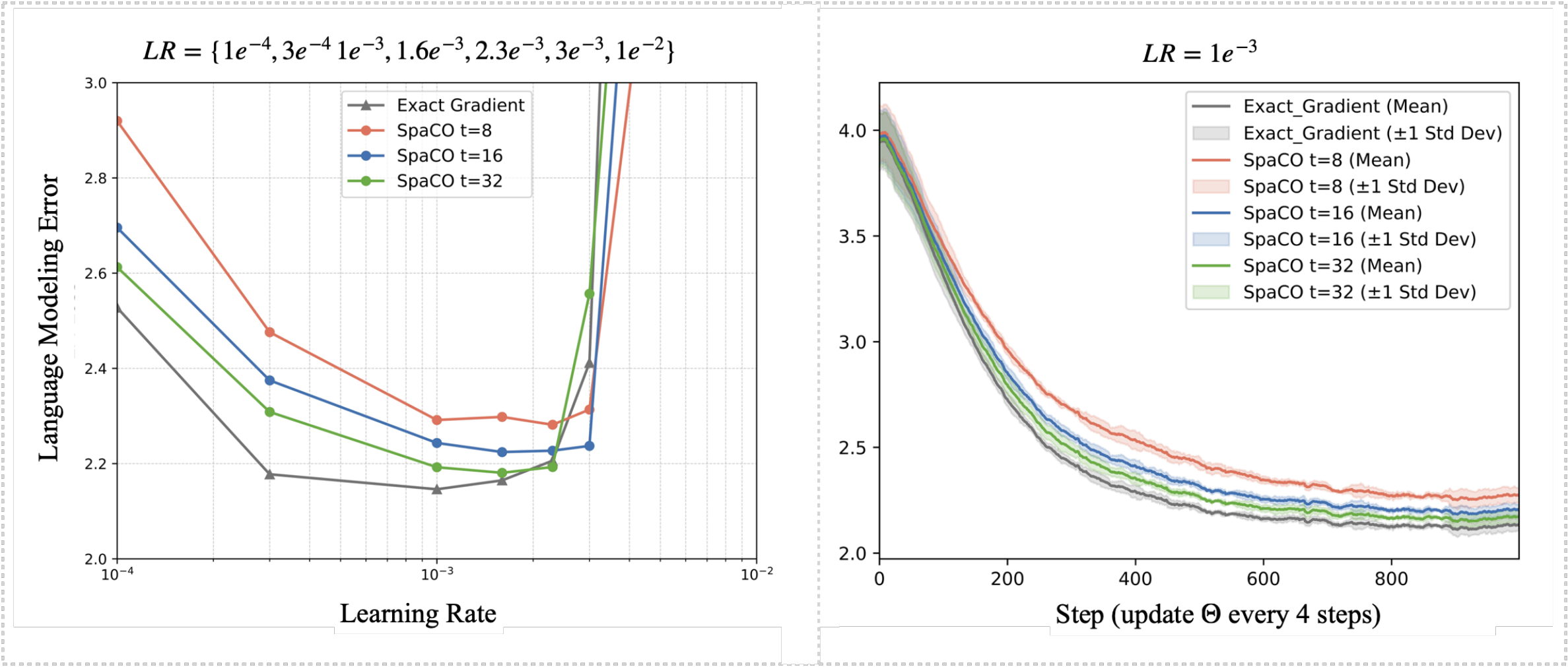}
\caption{(\textit{Left}) Performance comparison of SpaCO across $t=\{8,16,32\}$ under varying learning rates, using the same training setup and random seed.
(\textit{Right}) EMA-smoothed learning curves ($\alpha$=0.95) for SpaCO and exact gradient method, both trained with the optimal learning rate of 1e-3. 
}
\label{fig:lr}
\end{figure*}
\subsection{Effectiveness Analysis}
While SpaCO theoretically ensures unbiased gradient estimation through compensation factors, its increased gradient estimation variance raises practical effectiveness concerns.
To investigate this, we evaluate its performance using a common training recipe for context window extension: As outlined in the experimental setup, we extend LLaMA3-8B's original 8K window to 16K via LoRA.

For SpaCO, we fix the chunk size to 128 and conduct experiments with budgets of $t=\{8,16,32\}$, corresponding to sparsity ratios of $1/16$, $1/8$ and $1/4$ respectively. The training results using model parallelism combined with gradient checkpointing serves as the baseline reference. 
All training runs use a batch size of 4, allowing for a total of 250 parameter updates.
To mitigate numerical instability and gradient vanishing or explosion, we limit the compensation factor to a maximum value of 2.\footnote{Even in the absence of the compensation factor, excessively long gradient chains often result in vanishing or exploding gradients, diminishing their overall impact.}

\vspace{0.3\baselineskip}
\noindent
\textbf{Performance Under Varying Learning Rates.}
SpaCO introduces additional noises to the training process, necessitating independent tuning of the learning rate.
To this end, we perform grid search over seven learning rates \{1e-4,3e-4,1e-3,1.6e-3,2.3e-3,3e-3,1e-2\} to identify the optimal values.
Results are shown in Figure\,\ref{fig:lr}\,(\textit{Left}).

\vspace{0.3\baselineskip}
\noindent
\textbf{Learning Curves.}
We further record the learning curves for each configuration using the optimal LR of 1e-3, as determined from Figure\,\ref{fig:lr} (\textit{Left}). Training is conducted across four random seeds (controlling dataset shuffling), and we record the mean trajectories along with ±1 standard deviation bands. Results are presented in Figure\,\ref{fig:lr}\,(\textit{Right}).
The results demonstrate that with proper hyperparameter tuning, SpaCO achieves comparable performance to exact gradient training.

\vspace{0.3\baselineskip}
\noindent
\textbf{Practical Guidance.}
We recommend setting an upper bound (\emph{e.g.}, 2) on the compensation factor to reduce gradient estimation variance.
While omitting this constraint does not compromise training stability, it may lead to sub-optimal results.
Furthermore, although our experiments used fixed batch size and training iterations for fair comparisons, we suggest using larger batch size and more training epochs than exact gradient training.
This adjustment can promote better results in practical applications.

\section{Conclusion}
To address the critical challenge of efficiency in long-context LLM training, we introduce two training paradigms: SeCO and its enhanced variant SpaCO.
By partitioning the input sequence into smaller, manageable chunks and performing localized backpropagation for each chunk, SeCO achieves substantial memory savings.
Building upon this foundation, SpaCO introduces a carefully designed sparsification mechanism that randomly selects few chunks for backpropagation, reducing computational overhead.
The integration of a mathematically-grounded compensation factor ensures unbiased gradient estimation.
Our methods achieve impressive memory efficiency, enabling the fine-tuning of 8B models with 16K tokens on a single RTX 3090 GPU.
This represents a 16× memory reduction compared to naive parallel training.
SeCO and SpaCO significantly lower the barrier for practitioners working with long-context LLMs.

\section*{Limitations}
SeCO and SpaCO each present unique advantages but also have exhibit distinct limitations.
SeCO achieves accurate gradient computation and efficient memory usage but suffers from a quadratic increase in computation with sequence length, making it impractical for training on ultra-long sequences. In contrast, SpaCO significantly reduces computational cost and maintains comparable memory efficiency but sacrifices gradient accuracy, introducing substantial randomness that complicates convergence.
Ultimately, no single training strategy perfectly balances the trade-offs in all training scenarios. A practical approach requires identifying an optimal balance among the ``impossible triangle'' of computation, memory efficiency, and gradient accuracy.

\section*{Ethics Statement}
By optimizing memory consumption and computational efficiency, our approach not only lowers the financial barriers to training such models but also reduces energy consumption, contributing to more sustainable AI practices.

However, as with any significant technological advancement, ethical concerns must be considered. Lowering the cost and resource requirements for training long-context models may inadvertently enable the misuse of these models, including the creation of harmful or malicious language systems. It is essential to address these risks through responsible research practices and the development of robust safeguards.

\bibliography{custom}

\appendix

\begin{table}[h]
\centering
\caption{Arguments for DeepSpeed ZeRO3 offload}
\label{tab:zero3}
\begin{tabular}{lc}
\toprule 
\textbf{Argument} & \textbf{Value}\tabularnewline
\midrule 
overlap\_comm & true\tabularnewline
contiguous\_gradients & true\tabularnewline
reduce\_bucket\_size & 5e8\tabularnewline
stage3\_max\_live\_parameters & 1e9\tabularnewline
stage3\_max\_reuse\_distance & 1e9\tabularnewline
stage3\_prefetch\_bucket\_size & 5e8\tabularnewline
\bottomrule
\end{tabular}
\end{table}
\section{Experimental Datasets}
\textbf{PG19 Dataset.}
The PG19 corpus, an open-source long-text dataset released by DeepMind, is derived from books in the \href{https://www.gutenberg.org}{Project Gutenberg} repository published prior to 1919. This collection is supplemented with metadata containing book titles and publication dates.
For model training, we randomly selected 1,000 samples from the PG19 training partition. To ensure consistent sequence lengths, text samples exceeding 16K tokens were truncated to this threshold. The PG19 dataset is publicly available under the Apache License 2.0.

\section{Language Models}
\textbf{LLaMA3-8B.}
The LLaMA3-8B model, an open-source large language model developed by Meta AI, serves as the foundational model in our experiments. This selection is motivated by its widespread adoption within the research community. The licensing terms for the LLaMA3 series models are governed by the \href{https://github.com/meta-llama/llama3/blob/main/LICENSE}{Meta Llama 3 Community License Agreement}, which notably permits academic and commercial use with specific attribution requirements.

\section{Implementation Details}
\subsection{Pseudocode}
\label{app:pseudo-code}
The workflows of SeCO and SpaCO primarily manage the KV cache, focusing on its updates and the relay of gradients during backpropagation. These operations require overriding the default backpropagation mechanism in deep learning frameworks, which poses implementation challenges. To clarify this process, we provide pseudocode below.

\begin{table}[h]
\caption{Training results of SeCO vs. Model Parallelism (Baseline) across different learning rates.}
\label{tab:compare}
\centering
\begin{tabular}{cccc}
\toprule 
\multirow{2}{*}{\textbf{Method}} & \multicolumn{3}{c}{\textbf{LR}}\tabularnewline
\cmidrule{2-4}
 & 1e-4 & 3e-4 & 1e-3\tabularnewline
\midrule 
Baseline & 2.52 & 2.16 & 2.13\tabularnewline
SeCO & 2.53 & 2.18 & 2.15\tabularnewline
\bottomrule
\end{tabular}
\end{table}
\subsection{ZeRO3 Offload}
Detailed configurations are provided in Table\,\ref{tab:zero3}.
\label{app:zero3}

\section{Additional Results}
\label{app:more}
\textbf{Direct Validation of Gradient Accuracy.}
To assess the accuracy of the computed gradients, we conducted experiments using Qwen2.5-0.5B with \texttt{float64} precision. Gradients were obtained for sequences of 512 tokens using both naive parallel training and SeCO (with a chunk size of 64) and then compared element-wise. The results show that the gradients computed with SeCO achieve a precision exceeding 12 decimal places. The test code for this experiment is publicly available in our repository under the \texttt{test\_estimate} directory.

\vspace{0.3\baselineskip}
\noindent
\textbf{Indirect Validation of Gradient Accuracy.}
To evaluate SeCO’s performance in real training scenarios, we follow the experimental setup described in the main text. We compare SeCO’s training results with those obtained using model parallelism and gradient checkpointing. The results are summarized in Table\,\ref{tab:compare}.

The minor performance gap may be attributed to numerical issues arising from the increased number of operations in SeCO.
For example, FlashAttention introduces randomness during backpropagation due to the use of atomic additions (see\,\,\href{https://github.com/Dao-AILab/flash-attention/issues/414}{Github issue}).
Since SeCO involves tens of times more such operations than parallel training, it exhibits greater numerical instability.

\begin{lstlisting}[style = python]
def update_kv_cache(kv_cache, keys, vals):
    try:
        return concat(kv_cache.keys, keys), concat(kv_cache.vals, vals)
    finally:
        if is_gradient_enabled():
            kv_cache.keys.append(keys)
            kv_cache.vals.append(vals)
        else:
            k_detach, v_detach = keys.detach(), vals.detach()
            k_detach.requires_grad_(), v_detach.requires_grad_()
            kv_cache.keys.append(k_detach)
            kv_cache.vals.append(v_detach)

def grad_hook(grad, base, scaler=1):
    return grad + base * scaler

def copy_grad(a, b):
    for ak, av, bk, bv in zip(a.keys, a.vals, b.keys, b.vals):
        bk.register_hook(partial(grad_hook, base=ak.grad))
        bv.register_hook(partial(grad_hook, base=av.grad))
\end{lstlisting}

\end{document}